\pdfoutput=1

\documentclass[11pt]{article}

\usepackage[review]{acl}

\usepackage{times}
\usepackage{latexsym}
\usepackage{graphicx}
\usepackage{booktabs} 
\usepackage{amsmath} 
\usepackage{amssymb}
\usepackage[T1]{fontenc}
\usepackage{lipsum}

\usepackage[utf8]{inputenc}

\usepackage{microtype}

\title{Brain-Inspired Two-Stage Approach: Enhancing Mathematical Reasoning by Imitating Human Thought Processes}

\newcommand\blfootnote[1]{%
  \begingroup
  \renewcommand\thefootnote{}\footnote{#1}%
  \addtocounter{footnote}{-1}%
  \endgroup
}

\author{
    Yezeng Chen$^\Diamond$\textsuperscript{\rm 1,\rm2}, 
    Zui Chen$^\Diamond$\textsuperscript{\rm 1,\rm2}, 
    Yi Zhou$^\clubsuit$\textsuperscript{\rm 3} \\
    \textsuperscript{\rm 1}School of Information Science and Technology, ShanghaiTech University \\
    \textsuperscript{\rm 2}Shanghai Innovation Center for Processor Technologies \\
    \textsuperscript{\rm 3}School of Information Science and Technology, University of Science and Technology of China  \\
    \texttt{\{chenyz2022, chenzui2022\}@shanghaitech.edu.cn;} \\
    \texttt{yi\_zhou@ustc.edu.cn} 
}

\begin{document}
\maketitle

\blfootnote{$^\Diamond$ Equal Contribution.}
\blfootnote{$^\clubsuit$ Corresponding Authors.}

\begin{abstract}


Although large language models demonstrate emergent abilities in solving math word problems, there is a challenging task in complex multi-step mathematical reasoning tasks. To improve model performance on mathematical reasoning tasks, previous work has conducted supervised fine-tuning on open-source models by improving the quality and quantity of data. In this paper, we propose a novel approach, named Brain, to imitate human thought processes to enhance mathematical reasoning abilities, using the Frontal Lobe Model to generate plans, and then employing the Parietal Lobe Model to generate code and execute to obtain answers. First, we achieve SOTA performance in comparison with Code LLaMA 7B based models through this method. Secondly, we find that plans can be explicitly extracted from natural language, code, or formal language. Our code and data are publicly available at \url{https://github.com/cyzhh/Brain}.

\end{abstract}

\section{Introduction}


Although Large Language Models (LLMs) possess emergent abilities, including a certain level of Math Word Problem-solving ability in mathematical reasoning, whether through pre-training \citep{touvronLLaMAOpenEfficient2023, roziereCodeLlamaOpen2023, openaiGPT4TechnicalReport2023, anilPaLMTechnicalReport2023}, few-shot learning with prompts \citep{zhangCR2023, zhengPHP2023, zhuRules2023, wangPlanandSolvePromptingImproving2023}, through fine-tuning \citep{wangMathCoderSeamlessCode2023,yuanSCALINGRELATIONSHIPLEARNING2023,luoWizardMath2023} or verification \citep{dengUnifiedViewAnswer2023, wuGetMathProgressive2023,wangMathShepherdLabelFreeStepbyStep2023,romera-paredesMathematicalDiscoveriesProgram2023}, they lack strong logical reasoning skills and face challenges in complex multi-step mathematical reasoning tasks.

Previous research has explored various methods to expand the abilities of LLMs in complex multi-step mathematical reasoning tasks. If we consider a LLM as a human brain, then pre-training is akin to enhancing the general cognitive abilities of the entire brain. Fine-tuning and prompting  are analogous to optimizing the hippocampus, reinforcing the forms of knowledge input and memory. Verification is like enhancing the ability of the frontal and parietal lobes, consolidating and refining learned knowledge through testing and error correction.

Recent works \citep{yueMAmmoTHBuildingMath2023, gouToRAToolIntegratedReasoning2023, yuMetaMathBootstrapYour2023} attempt to enhance the ability of LLMs in complex multi-step mathematical reasoning tasks by increasing the amount and improving the quality of supervised fine-tuning (SFT) training data, we believe that simply activating the hippocampus of the LLMs is not enough. It is necessary to activate the corresponding ability in each brain region. Specifically, by activating the frontal and parietal lobes respectively, we can correspondingly activate the ability to understand problem-solving strategies and to understand coding, thereby improving the ability of LLMs in complex multi-step mathematical reasoning tasks.

\begin{figure*}[ht]
  \centering
  \includegraphics[page=1,width=0.97\linewidth,trim=0 0 0 0,clip]{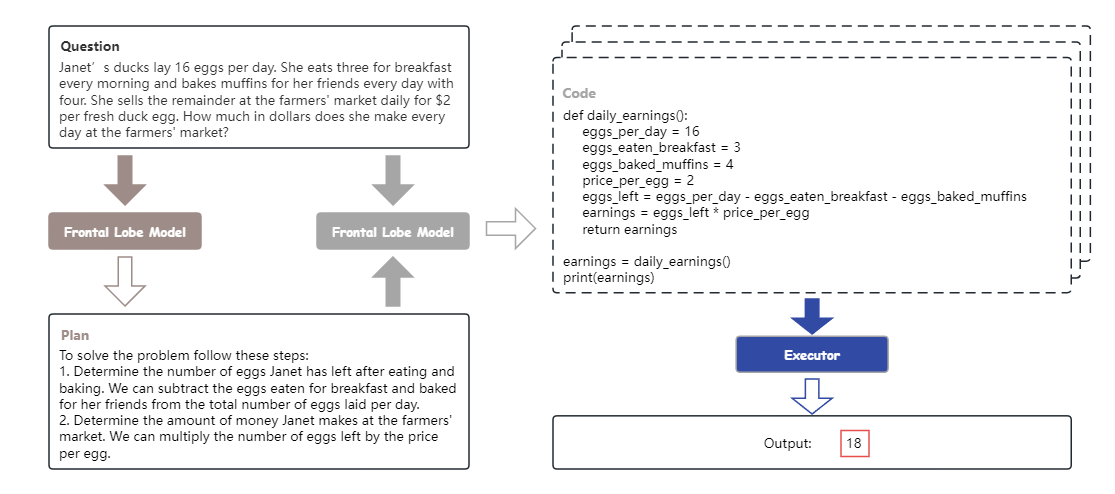}
  \caption{Brain, using a combined approach of the Frontal Lobe Model and the Parietal Lobe Model to simulate the human problem-solving thought process.}
  \label{fig:Brain}
\end{figure*}

\begin{figure*}[ht]
  \centering
  \includegraphics[page=1,width=0.97\linewidth,trim=0 0 0 0,clip]{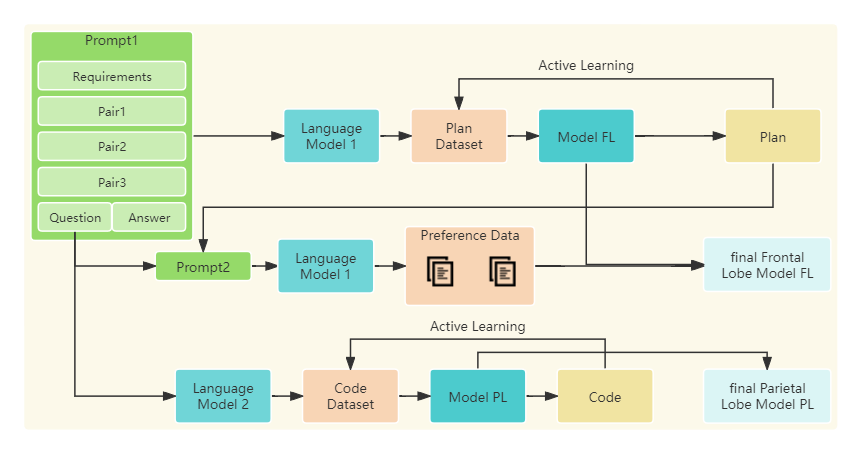}
  \caption{The overview of our proposed method, Brain.}
  \label{fig:method}
\end{figure*}

Process Reward Model (PRM) \citep{zhuSolvingMathWord2023, lightmanLetVerifyStep2023, wangMathShepherdLabelFreeStepbyStep2023} and Lean Reward Model (LRM) can significantly reduce the occurrence of erroneous steps and thus enhance the performance of LLMs. PRM evaluates the reasoning paths step-by-step, while LRM enables the model to transform its generated Chain of Thought (CoT) process into a Lean \footnote{https://lean-lang.org/} format, then assesses the correctness of the process through the Lean calculation results. The manual annotation required for PRM, especially for complex multi-step reasoning tasks, demands high annotator skills and is costly, with LRM being even more expensive. Additionally, these two types of validation models all train the "frontal lobe" and "parietal lobe" regions of LLMs simultaneously, meaning the LLM cannot concurrently have the abilities of both planning and reasoning calculation effectively.

However, we believe that the logical abilities of LLMs in mathematical reasoning have not been fully demonstrated, as LLMs cannot simultaneously manage both planning and reasoning calculation. In this paper, we aim to address the following research questions: \textbf{($\spadesuit$1)} Does the output of LLMs contain a plan? If so, can it be explicitly extracted? \textbf{($\spadesuit$2)} How can we use the plan to solve complex reasoning tasks? \textbf{($\spadesuit$3)} How can we construct higher-quality plans? Is a high-quality plan necessarily useful?

To address this challenge and overcome the limitation of LLMs' ability, in this paper, we propose a two-stage technique called \textbf{Brain}, whose overview is shown in Figure \ref{fig:Brain}. It involves constructing a novel step-level model framework for solving mathematical reasoning tasks to simulate the human approach to coding for solving real-world issues: 1) Inspired by the use of high-level directives and meta-prompting to guide the LM in breaking down complex tasks into smaller, more manageable sub-tasks \citep{hongMetaGPTMetaProgramming2023, suzgunMetaPromptingEnhancingLanguage2024}, we decompose complex mathematical reasoning problems into two steps using the same LLM: planning based on the question, followed by code generation based on the plan. 2) These two tasks are assigned to specialized models for the Frontal Lobe Model for decision-making and the Parietal Lobe Model for code structure and logical flow. We use Direct Preference Optimization (DPO) to optimize the Models to automatically select plans that align more closely with the question. 


Overall, in this paper, our contributions are as follows: 

\begin{itemize}
    \item We propose a novel approach Brain that imitate human brain thought processes to enhance mathematical reasoning ablilities.
    \item Brain achieves SOTA performance in comparison with Code LLaMA 7B based models with zero-shot.
    \item We find that plans can be explicitly extracted from natural language, code, or formal language.
\end{itemize}

\section{Related Work}


In this section, we provide a brief overview of the progress made in mathematical reasoning tasks, emphasizing their relevance and connection to our work.


\textbf{Reasoning Format.} 
Recent works have utilized natural language \citep{yu2023outcomesupervised, zhangCR2023,zhuRules2023}, code \citep{wuGetMathProgressive2023, yueMAmmoTHBuildingMath2023,wangMathCoderSeamlessCode2023,gouToRAToolIntegratedReasoning2023}, and formal language \citep{gaoGLLaVASolvingGeometric2023, trinhSolvingOlympiadGeometry2024} to solve complex mathematical reasoning tasks. A two-stage training method \citep{shaoDeepSeekMathPushingLimits2024} has demonstrated that code-based training is beneficial for program-assisted mathematical reasoning, thereby directly leveraging the contextual learning abilities of LLMs to generate more precise and rigorous deductive reasoning.

\textbf{Planning.} Despite significant progress in enhancing the capability of LLMs to handle complex reasoning tasks through appropriate prompting for global planning \citep{yaoTreeThoughtsDeliberate2023}, construction of step-by-step reasoning chains \citep{wangMathShepherdLabelFreeStepbyStep2023,khotDecomposedPromptingModular2023}, and the use of external tools \citep{gouToRAToolIntegratedReasoning2023,yueMAmmoTHBuildingMath2023, shaoDeepSeekMathPushingLimits2024}, these methods still have some limitations. Recent research \citep{suzgunMetaPromptingEnhancingLanguage2024} has made strides by using high-level directives and meta-prompting to guide LMs in breaking down complex tasks into smaller, more manageable sub-tasks. However, these methods not only require more resources and greater time complexity but also cannot fully activate the large model's abilities in mathematical reasoning.

\textbf{Self-Verification.} Some methods \citep{yu2023outcomesupervised, dengUnifiedViewAnswer2023} calibrate through answers is insufficient for LLMs to generate high-quality reasoning steps and answers. Therefore, we need to focus on fine-grained step-level self-verification \citep{wuFineGrainedHumanFeedback2023}. Current research focuses on reverse verification by masking numbers in reasoning paths \citep{wuGetMathProgressive2023, liMakingLargeLanguage}, adding realism to predictions by post-editing reasoning chains based on external knowledge or tools \citep{zhaoVerifyandEditKnowledgeEnhancedChainofThought2023,sunAdaPlannerAdaptivePlanning2023}, modifying reasoning steps without external feedback \citep{hongCloserLookSelfVerification2023,huangLargeLanguageModels2023} and decomposing reasoning steps and self-correct each step \citep{paulREFINERReasoningFeedback2023, lingDeductiveVerificationChainofThought2023}. Also, research has been conducted to optimize self-verification performance through Process Reward Model (PRM), including utilizing human annotators \citep{zhuSolvingMathWord2023,lightmanLetVerifyStep2023}, Monte Carlo Tree Search \citep{wangMathShepherdLabelFreeStepbyStep2023,haoReasoningLanguageModel2023}, and InternLM2-Math \footnote{https://github.com/InternLM/InternLM-Math}, which introduces Outcome Reward Model (ORM), Process Reward Model (PRM), and Lean as Reward Model (LRM) simultaneously. It is demonstrated that more fine-grained internal and external feedback can significantly improve the accuracy, quality, and stability of model outputs.


\section{Brain}

\subsection{Overview}

Figure \ref{fig:method} shows the overview of training Brain. In this section, we want to emphasize three key points: 
\begin{itemize}
    \item[1.] Utilizing prompts to obtain a large number of high-quality plan datasets and preference datasets from LLMs (\S \ref{subsec:3.2});
    \item[2.] Training the frontal lobe model to better generate high-quality Plans from problems (\S \ref{subsec:3.3});
    \item[3.] Training the parietal lobe model to better generate high-quality code-form reasoning paths from Plans, ultimately achieving highly accurate answers (\S \ref{subsec:3.4}).
\end{itemize}

\begin{figure*}[h]
\centering
\fbox{%
  \begin{minipage}{0.9\textwidth}
    \textcolor{gray}{\textbf{The few-shot prompt $C$ for GSM8K high-quality plan creating.}} \\
    \textcolor{black}{\textbf{R:}} \textcolor{black}{Let's write a plan based on the python program for solving math problems.} \\
    \textcolor{teal}{\textbf{Question:}} \\
    \textcolor{teal}{Richard lives in an apartment building with 15 floors. Each floor contains 8 units, and 3/4 of the building is occupied. What's the total number of unoccupied units In the building?} \\
    \textcolor{teal}{\textbf{Program:}} \\
    \textcolor{teal}{\char39\char39\char39 python} \\
    \textcolor{teal}{def solution():} \\
    \textcolor{teal}{\qquad num\_floors = 15} \\
    \textcolor{teal}{\qquad num\_units\_per\_floor = 8} \\
    \textcolor{teal}{\qquad num\_total\_floors = num\_floors * num\_units\_per\_floor} \\
    \textcolor{teal}{\qquad num\_units\_occupied = num\_floors * num\_units\_per\_floor * 0.75} \\
    \textcolor{teal}{\qquad num\_units\_unoccupied = num\_total\_floors - num\_units\_occupied} \\
    \textcolor{teal}{\qquad return num\_units\_unoccupied} \\
    \textcolor{teal}{\char39\char39\char39 } \\
    \textcolor{blue}{\textbf{Plan:}} \\
    \textcolor{blue}{To solve the problem follow these steps:} \\
    \textcolor{blue}{1. Determine the total number of units in the building. We can multiply the number of units per floor by the total number of floors.} \\
    \textcolor{blue}{2. Determine the number of occupied units. We can multiply the total number of units by the fraction representing the occupied units.} \\
    \textcolor{blue}{3. Determine the number of unoccupied units. We can subtract the number of occupied units from the total number of units.} \\
    \textcolor{black}{...} 
  \end{minipage}
}
\captionsetup{type=figure}
\caption{The few-shot prompt $C$for GSM8K high-quality plan creating. The black color text is the requirement $R$. The teal and blue color text is the one of example pairs, which contains input $\overline{x_1}$ including example question and program and the output $\overline{y_1}$ including example plan.}
\label{fig:C_plan}
\end{figure*}

\subsection{Prompting}
\label{subsec:3.2}

We obtained a large number of high-quality plan datasets and preference datasets from the large model \textit{gpt-3.5-turbo-1106} using prompts, based on the following observations:
\begin{itemize}
    \item To mimic the human brain's problem-solving approach, we require a substantial amount of high-quality datasets to supervised fine-tune and direct preference optimize the Frontal Lobe Model and the Parietal Lobe Model. Using the model \textit{gpt-3.5-turbo-1106} can generate datasets of fairly high quality while ensuring minimal costs;
    \item Structuring the inputs and outputs of the Models facilitates easier parsing the plan datasets and the preference dataset that we create.
\end{itemize}


We design different few-shot prompts for various steps as outlined in Figure \ref{fig:method}. The prompts frequently used in Brain are shown in Figure \ref{fig:C_plan} and \ref{fig:C_score}, and all other prompts used in the experiments will be presented in the Appendix \ref{prompt}. 

We provide prompt $C$, along with the corresponding task input $x$, and generate output $y$ from GPT:
\begin{equation}
    \mathcal{G}(y|C,x) = \prod_{t=1}^{|y|}\mathcal{G}_{gpt}(y_t|C,x,y_{<t}),
\end{equation}


where $C$ integrates three completely different human-annotated example pairs $\mathrm{Pair}_i$ and the customized requirement $R$ for the tasks:
\begin{equation}
    \mathrm{Pair}_i = (\overline{x_i},\overline{y_i}),
\end{equation}
\begin{equation}
    C = R \oplus \mathrm{Pair}_1 \oplus \mathrm{Pair}_2 \oplus \mathrm{Pair}_3, \\
\end{equation}


Based on two different tasks: generating high-quality plan datasets and high-quality preference datasets, we designed two different few-shot prompts $C$ shown in Figure \ref{fig:C_plan} and $C^{'}$ shown in Figure \ref{fig:C_score}. 


In Prompt $C$, the input $x_i$ includes the question and solution, and the output $y_i$ represents the plan. Meanwhile, in prompt $C^{'}$, the input $x_i$ includes the question and plan, and the output $y_i$ represents the score and reasons.

\begin{figure*}
\centering
\fbox{%
  \begin{minipage}{0.9\textwidth}
    \textcolor{gray}{\textbf{The few-shot prompt $C^{'}$ for GSM8K high-quality score dataset creating}} \\
    \textcolor{black}{\textbf{R:}} \textcolor{black}{Review the following Question, Plan and the corresponding response Solution using the additive 3-pointscoring system described below. Points are accumulated based on the satisfaction of eachcriterion:} \\
    \textcolor{black}{1. Points are awarded according to how well the logic of the Plan is aligned with the Question.} \\
    \textcolor{black}{2. Points are awarded according to how the Plan is non-redundant and non-duplicate.} \\ 
    \textcolor{black}{3. Points are awarded according to how the necessary steps and omit key details or steps.} \\
    \textcolor{black}{Based on the given Question and Plan, briefly justify your total score. Conclude with the score using the format: Score: [<total points>].Remember to assess from the AI Assistant perspective, and evaluate the response in alignment with this additive scoring model, we'll systematically attribute points based on the outlined criteria.} \\
    \textcolor{teal}{\textbf{Question:}} \\
    \textcolor{teal}{Josh decides to try flipping a house.  He buys a house for \$80,000 and then puts in \$50,000 in repairs.  This increased the value of the house by 150\%.  How much profit did he make?} \\
    \textcolor{teal}{\textbf{Plan:}} \\
    \textcolor{teal}{1. Determine the value of the house before the repairs. We can subtract the cost of the repairs from the original price.} \\
    \textcolor{teal}{2. Determine the value of the house after the repairs. We can multiply the value of the house before the repairs by 150\%.} \\
    \textcolor{teal}{3. Determine the profit. We can subtract the cost of the repairs from the value of the house after the repairs.} \\
    \textcolor{blue}{\textbf{Solution:}} \\
    \textcolor{blue}{1. Alignment of Plan with the Question:} \\
    \textcolor{blue}{The plan starts by determining the value of the house before the repairs, which is unnecessary since the question already states that Josh bought the house for \$80,000. This step does not align well with the question.} \\
    \textcolor{blue}{The second step in the plan suggests multiplying the value of the house before the repairs by 150\%. This is a misinterpretation of the question. The question states that the value of the house increased by 150\% due to repairs, not that its final value is 150\% of the original price.}\\
    \textcolor{blue}{The third step, determining the profit by subtracting the cost of repairs from the value of the house after repairs, aligns with the question's requirement but is based on an incorrectly calculated value of the house after repairs.} \\
    \textcolor{blue}{Score for this criterion: 0} \\
    \textcolor{blue}{2. Non-redundancy and Non-duplication:} \\
    \textcolor{blue}{The first step in the plan is redundant as the initial value of the house is already given in the question.} \\
    \textcolor{blue}{The other steps are not redundant but are based on an incorrect understanding of the question.} \\
    \textcolor{blue}{Score for this criterion: 0.5} \\
    \textcolor{blue}{3. Completeness and Detail:} \\
    \textcolor{blue}{The plan fails to correctly outline the necessary steps for solving the problem. The misinterpretation of how the 150\% increase in value applies significantly impacss and accuracy of the plan.} \\
    \textcolor{blue}{The plan does cover tts the completenehe concept of calculating profit, which is a required step, but it is based on incorrect prior steps.} \\
    \textcolor{blue}{Score for this criterion: 0} \\
    \textcolor{blue}{Based on these assessments, the total score is calculated as the sum of the points from each criterion. Score: [0 + 0.5 + 0] = [0.5]} \\
    \textcolor{black}{...} 
  \end{minipage}
}
\captionsetup{type=figure}
\caption{The few-shot prompt $C^{'}$ for GSM8K high-quality score dataset creating. The black color text is the requirement $R$. The teal and blue color text is the one of example pairs, which contains input $\overline{x_1}$ including example question and example plan $\overline{z_1}$ and the output $\overline{y_1}$ including example reasons and score.}
\label{fig:C_score}
\end{figure*}

\subsection{Frontal Lobe Model}
\label{subsec:3.3}


In this subsection, we introduce a method for training the Frontal Lobe Model, enabling it to better generate high-quality Plans from problems. This ensures an improvement in accuracy when the parietal lobe model performs reasoning.


To address the issue of LLMs generating Plans of low quality when solving complex reasoning tasks, we adopt a straightforward method to optimize fine-tuned model, known as Direct Preference Optimization (DPO) \citep{rafailovDirectPreferenceOptimization2023}, which allows for the extraction of its optimal policy in closed form without the need for a reinforcement learning training loop.


\textbf{Datasets.} We utilized four models from the ToRA series: ToRA-Code 7B, ToRA-Code 13B, ToRA-Code 34B, and ToRA 70B, to perform 100 inference samplings on the GSM8K train set, resulting in 3,000K reasoning paths. Then, we deduplicated and filtered all the obtained reasoning paths, ultimately obtaining 90K distinct and correct reasoning paths dataset $D$.


To enhance the model's generalization ability and performance within the domain, it is crucial to ensure that the source dataset for SFT and DPO are different. From the dataset $D$ comprising 7.5K questions, we extracted all reasoning paths of 5K questions as the source dataset $D_{sft}$ for creating the SFT dataset and all reasoning paths of 2.5K questions as the source dataset $D_{dpo}$ for creating the preference dataset.


\textbf{Supervised Fine-Tuning.} To obtain high-quality plans, we utilize prompt $C$ with the GPT model \textit{gpt-3.5-turbo-1106} to perform one round of inference on the source data $D_{sft}$, generating the plan dataset $P_{sft}$. Subsequently, we conduct SFT on the dataset $P_{sft}$ using Code LLaMA 7B, resulting in the training of the Frontal Lobe Model $\mathcal{FL}_0$.


Based on $\mathcal{FL}_0$, we conducted active learning for k times, stopping when the performance of the model $\mathcal{FL}_i$ on the GSM8K test set no longer improves after iterations, to serve as our initial version of the model $\mathcal{FL}$. We use the plans generated by $\mathcal{FL}_i$ for inference in the Parietal Lobe Model, which will be introduced in $\S \ref{subsec:3.4}$, to evaluate the performance of the Frontal Lobe Model.


\textbf{Direct Preference Optimization.} We verify and optimize our current model's ability to generate plans without using reinforcement learning, by employing Directed Preference Optimization. We use whether the logic of the plan aligns with the question, whether steps in the plan are repeated, and whether the plan omits key steps as the scoring criteria $R^{'}$. The input $x_i$ includes the question and the plan $z_i$ generated by the Frontal Lobe Model $\mathcal{FL}$, and the manually annotated scores and reasons form a new Pair, which serves as our prompt $C^{'}$, as shown in Figure \ref{fig:C_score}. 
\begin{equation}
    C^{'} = R^{'} \oplus \mathrm{Pair}^{'}_1 \oplus \mathrm{Pair}^{'}_2 \oplus \mathrm{Pair}^{'}_3, \\
\end{equation}


After generating plans $z$ through inference, we use the GPT to generate scores $s$, which means how well the plans align with the question:
\begin{equation}
    \mathcal{G}(y|C^{'},x) = \mathcal{G}_{\mathcal{FL}}(z|C,x) \cdot \mathcal{G}_{gpt}(y|C^{'},x,z),
\end{equation}
\begin{equation}
    score = \mathcal{E}(y).
\end{equation}


We use prompt $C^{'}$ on the model \textit{gpt-3.5-turbo-1106} to score the preferences of the generated plan datasets and provide reasons, extracting the scores to obtain the preference dataset.


Then, we perform DPO on the initial version of the frontal lobe model using the constructed preference dataset , resulting in the optimized Frontal Lobe Model $\mathcal{FL}^{*}$.

\subsection{Parietal Lobe Model}
\label{subsec:3.4}

In this subsection, we introduce a method for training the parietal lobe model, enabling it to better generate high-quality code-form reasoning paths from Plans, with the aim of achieving higher accuracy.

\textbf{Datasets.} We use the model \textit{gpt-3.5-turbo-1106} to perform inference on Dataset $D$, with the generated plans and questions as input and the code-form reasoning paths and answers as output, to serve as our SFT dataset $Q_0$ for training the Parietal Lobe Model. Otherwise, We performed 100 inference samplings on the GSM8K train set using model $\mathcal{PL}$ and processed the obtained reasoning paths to remove duplicates and filter out incorrect reasoning paths, resulting in dataset $Q_i$, where $i$ means iteration times.

\textbf{Supervised Fine-Tuning.} We conducted supervised fine-tuning on the obtained Dataset $Q_0$ using Code LLaMA 7B, resulting in the first version of the Parietal Lobe Model, $\mathcal{PL}_0$.
Based on $\mathcal{PL}_0$, we conducted active learning for k times, stopping when the performance of model $\mathcal{PL}_i$ on the GSM8K test set no longer improves after iterations, to serve as our initial version of the model $\mathcal{PL}$. 

\section{Experiments}


\subsection{Experiment Settings}

\textbf{Models.} We use four ToRA language models: ToRA-CODE 7B/13B/34B and ToRA 70B with default parameters except a temperature of 0.9 in 100 sample times to create source reasoning paths dataset $D$. We use two OpenAI language models: \textit{gpt-3.5-turbo-1106} and \textit{gpt-4-1106-preview} with default parameters in sampling for creating plan datasets and preference dataset. And We use Code LLaMA 7B to train our model Brain.

\textbf{Datasets.} We conducted SFT with 55K data for $\mathcal{FL}$, 90K data for $\mathcal{PL}$, and 2.5K data for DPO. And we evaluated the models on GSM8K \citep{cobbeTrainingVerifiersSolve2021}.

\subsection{Main Results \& Analysis}


Table \ref{tab:result} shows the overall experiment results. We mainly compare our approach with LLMs of size 7B, using zero-shot inference. Experimental results show that, aside from models trained on exceptionally strong baseline models like InternLM2-Base and DeepSeekMath-Base, the performance of the model trained with our proposed novel method achieves SOTA performance in comparison with Code LLaMA 7B based models, with 74\% Accuracy.

\begin{table}[ht]
\centering
\begin{tabular}{lc}
\toprule
\textbf{Model} & \textbf{Accuracy}  \\
\midrule
{\small\textit{Closed-Source Models}} & \\
Minerva & 58.8\% \\
PaLM-2 & 80.7\% \\
GPT-3.5-turbo & 80.8\% \\
GPT-4 & 92.0\%  \\
\midrule
{\small\textit{Open-Source Models without Code}} & \\
LLaMA2 & 16.0 \% \\
Llemma & 36.4\% \\
InternLM2-Base & 36.5\% \\
InternLM2-Math-Base & 49.2\% \\
MAmmoTH & 53.6\% \\
MetaMath & 66.5\% \\
DeepSeekMath-Base & 64.2\% \\
\midrule
{\small\textit{Open-Source Models with Code}} & \\
Code LLaMA & 20.8\% \\
MAmmoTH-Coder & 59.4\% \\
MathCoder-CL & 67.8\%  \\
ToRA-Code & 72.6\% \\
InternLM2-Math & 78.1\% \\
DeepSeekMath-RL & 86.7\% \\
\textbf{Brain} & \textbf{74\%} \\
\bottomrule
\end{tabular}
\caption{Main Results on GSM8K. Comparison for Code LLaMA 7B based models with Zero-Shot.}
\label{tab:result}
\end{table}


\textbf{Language Models can extract plan explicitly.} To explore the best training effects on Code LLaMA 7B using plan datasets generated from the same source data, we adopted two methods to generate plans. For using Question and Code to generate Plan, we extracted the most frequently repeated paths from the previously generated dataset $P$ during the process of removing duplicate reasoning paths to form dataset $D^{'}$. Using prompt $C$, we conducted one sampling on both $D$ and $D^{'}$ on the large model \textit{gpt-3.5-turbo-1106} to obtain plan datasets $P$ and $P^{'}$ , respectively, and then performed inference on model $\mathcal{PL}$. For using Question and Solution to generate Plan, using prompt $C^{''}$, we conducted one sampling on the GSM8K train set with the model \textit{gpt-3.5-turbo-1106} to obtain the plan dataset $P^{''}$ , and similarly performed inference on model $\mathcal{PL}$. 


According to the experimental results in Table \ref{tab:Q+CvsQ+S}, it is revealed that the outputs of LLMs for mathematical reasoning tasks, whether in natural language, code, or formal language, all contain plans. We were able to explicitly extract these plans using different but similar prompts. Under the same data scale, the performance of LLMs fluctuates around 70\%, and filtering out incorrect data has almost no impact.

At the same time, it can be understood that using correct answers to prompt GPT to generate plans can maximize the automatic generation of high-quality plan datasets under the same data volume. However, expanding the dataset's size through 100 samplings enables the highest quality of plan datasets generated from the same source data.

\begin{table}[h]
\centering
\small
{%
\begin{tabular}{ccc}
\hline
\textbf{Dataset} & \textbf{Wrong case} & \textbf{Accuracy} \\
\hline
$P$ & $\checkmark$ & 73.7 \\
$P$ &  & 73.8 \\
$P^{'}$ & $\checkmark$ &  69.7 \\
$P^{'}$ &  &  70.1 \\
$P^{''}$ & $\checkmark$  & 71.0 \\
$P^{''}$ &  & 71.7 \\
\hline
\end{tabular}
}
\caption{Exploring the best training effects on Code LLaMA 7B using plan datasets generated from the same source data and compare for whether use wrong case }
\label{tab:Q+CvsQ+S}
\end{table}

\textbf{Better plan, better performance.} We use prompt $C^{''}$ with the model \textit{gpt-3.5-turbo-1106} as our Frontal Lobe Model to generate plans on the GSM8K test set, and then use our trained parietal lobe model $\mathcal{PL}$ to generate code and execute it to obtain answers. 

According to the experimental results in Table \ref{tab:gptvspl}, it is revealed that the two-stage framework of Brain, which simulates the human approach to problem-solving, is effective, achieving improvements of 2.9\% (69.0\% $\rightarrow$ 71.9\%) on Code LLaMA 7B and 1.0\% (80.8\% $\rightarrow$ 81.8\%) on \textit{gpt-3.5-turbo-1106}. Moreover, the higher the quality of plans generated in the first stage, the higher the accuracy in solving complex reasoning tasks, using the model \textit{gpt-3.5-turbo-1106} as the Frontal Lobe Model resulted in a 7.5\% improvement compared to $\mathcal{FL}^{*}$.

Based on the case study in Appendix \ref{casestudy}, we find that the accuracy of \textit{gpt-3.5-turbo-1106} and $\mathcal{PL}$ in tasks of aligning codes with plans is as high as 90\%. 

\begin{table}[h]
\centering
\small
{%
\begin{tabular}{lc}
\hline
\textbf{Method} & \textbf{Accuracy} \\
\hline
\textit{one-stage} & \\
SFT & 69.0\% \\
GPT-3.5-turbo & 80.8\% \\
\hline
\textit{Brain(two-stage)} & \\
$\mathcal{FL}$ + $\mathcal{PL}$ & 71.9\% \\
$\mathcal{FL}^{*}$ + $\mathcal{PL}$ & 72.9\% \\
$\mathcal{FL}^{*}_{all}$ + $\mathcal{PL}$ & 74.0\% \\
\textit{gpt-3.5-turbo-1106} + $\mathcal{PL}$ & 80.4\% \\ 
\textit{gpt-3.5-turbo-1106} + \textit{gpt-3.5-turbo-1106} & 81.8\% \\
\hline
\end{tabular}
}
\caption{Investigation of how two-stage method and how the quality plan affect model performance of complex math reasoning tasks.}
\label{tab:gptvspl}
\end{table}


We conducted one inference sampling on the GSM8K train set using the GPT models \textit{gpt-3.5-turbo-1106} and \textit{gpt-4-1106-preview}, respectively. The datasets obtained were then fine-tuned on Code LLaMA 7B, training the models the first version of the model $\mathcal{FL}_{gpt3.5}$ and $\mathcal{FL}_{gpt4}$. Inference was performed on the GSM8K test set with these models, generating datasets $P_{gpt3.5}$ and $P_{gpt4}$ . Inference was carried out on $PL$.

According to Table \ref{tab:gptvs}, although the performance of \textit{gpt-4-1106-preview} was slightly higher than that of \textit{gpt-3.5-turbo-1106} by 1.6\%, we chose to use \textit{gpt-3.5-turbo-1106} for the sake of consistency in the models used in our experiments and cost considerations, given the extensive reliance on GPT for subsequent experiments.

\begin{table}[h]
\centering
\small
{%
\begin{tabular}{lc}
\hline
\textbf{Method} & \textbf{Accuracy} \\
\hline
$\mathcal{FL}_{gpt3.5}$ + $\mathcal{PL}$ & 69.7 \\
$\mathcal{FL}_{gpt4}$ + $\mathcal{PL}$ & 71.5 \\
\hline
\end{tabular}
}
\caption{Investigating whether using GPT-4 to generate plan datasets will affect model performance of complex math reasoning tasks.}
\label{tab:gptvs}
\end{table}




\section{Conclusion}



We present Brain, which currently shows competitive performance on the GSM8K dataset compared to other open-source models. At the same time, we attempted to use Brain's framework on GPT, resulting in a significant improvement. Our extensive ablation experiments indicate that the outputs of LLMs for mathematical reasoning tasks, whether in natural language, code, or formal language, all contain plans. By introducing a two-stage framework that breaks down complex reasoning tasks into two steps, we train two models to simulate two regions of the human brain, thereby using plans to solve complex reasoning tasks. We used Direct Preference Optimization, a simple method that optimizes strategy directly using preferences, allowing for the extraction of its optimal policy in a closed form without the need for a reinforcement learning training loop. This approach generates high-quality plans, aiding LLMs in producing more accurate reasoning paths and answers.


Furthermore, we discovered that the degree of alignment between the plan and the question positively correlates with the alignment between the code and the plan. LLMs implicitly score the alignment of the plan with the question. If the score is low, the model will not generate code based on the plan but will instead regenerate the code. In the future, we will explore how LLMs follow plans to generate reasoning paths, to explain the error correction abilities of LLMs.

\newpage
\section*{Limitations}


The main limitation of this paper is that it does not analyze additional methods to enhance performance on Brain, such as self-consistency and self-verification. Furthermore, the Brain framework could also be applied to other open-source models. We should also evaluate complex mathematical reasoning tasks on a broader range of tasks to make the results more convincing.

\section*{Ethical Statements}

We claim from these aspects of ethical risks: 

1) Our work leverages the open-source model CodeLLaMA and the GSM8K dataset. We have strictly followed their licensing protocols to ensure compliance with all usage terms and conditions.

2) We utilized GPT4 for translation and grammatical corrections in our manuscript. All generated content has been thoroughly reviewed and revised by human authors to ensure it adheres to ethical guidelines and maintains the integrity of our research.

3) Our research involves the generation of plan datasets and preference datasets using GPT. While we have conducted sample checks to identify any ethical issues and found none, we recognize the limitations of this approach. It is not feasible to guarantee that all outputs generated by GPT are free from ethical risks.

\bibliography{anthology,custom}
\bibliographystyle{acl_natbib}

\appendix

\onecolumn
\section{Experiment Details}


The entire training process of Brain on NVIDIA A100 40G GPUs took a total of 8 hours and we evaluating GSM8K test set took an average of 3 minutes each time. Both SFT and active learning used a learning rate of 2e-5 with a 3\% warm-up period for 1 epoch and a global batch size of 128. For DPO, aside from the learning rate being 2e-6, all other parameters were set to default. We trained all models with DeepSpeed ZeRO Stage3 and Flash-Attention 2. Additionally, GSM8K has 7473 train set and 1319 test set. The experimental results are based on just a single run.

\section{Case Study}
\label{casestudy}


\begin{table*}[h]
\centering
\small
{%
\begin{tabular}{|l|cccccccccc|}
\hline
\textbf{id} & 0 & 1 & 2 & 3 & 4 & 5 & 6 & 7 & 8 & 9 \\
\hline
Plan Align Question($\mathcal{FL}$) & 1 & 1 & 0 & 1 & 1 & 1 & 1 & 0.5 & 0.5 & 1 \\
Code Align Plan($\mathcal{PL}$) & 1 & 1 & 0 & 1 & 1 & 1 & 1& 1 & 1 & 1 \\
Code Align Plan(\textit{gpt-3.5-turbo-1106}) & 1 & 1 & 0 & 1 & 1 & 1 & 1 & 1 & 1 & 1 \\ 
score & T & T & F & T & T & T & T & F & F & T \\
\hline
\end{tabular}
}
\caption{For the Plan Align Question, we score the 10 sampled examples based on the scoring rules for plans in the prompt, with the score being the average score for each step. For Code Align Plan, the scoring is based on the extent to which the generated code completely conforms to the steps, with the score being the average score for each step.}
\label{tab:casestudy}
\end{table*}

\section{Prompt}
\label{prompt}

\begin{figure*}[h]
\centering
\fbox{%
  \begin{minipage}{0.9\textwidth}
    \textcolor{gray}{\textbf{The few-shot prompt $C^{''}$ for GSM8K high-quality plan creating.}} \\
    \textcolor{black}{\textbf{R:}} \textcolor{black}{Let's write a plan based on the solution for solving math problems.} \\
    \textcolor{black}{\textbf{Question:}} \\
    \textcolor{black}{Richard lives in an apartment building with 15 floors. Each floor contains 8 units, and 3/4 of the building is occupied. What's the total number of unoccupied units In the building?} \\
    \textcolor{teal}{\textbf{Solution:}} \\
    \textcolor{teal}{1. The total number of units in the building will be 8 units/floor * 15 floors = <<8*15=120>>120 units.} \\
    \textcolor{teal}{2. If 3/4 of the building is occupied, then the total number of occupied units is 3/4 * 120 units = <<3/4*120=90>>90 units.} \\
    \textcolor{teal}{3. The total number of unoccupied units is 120 units - 90 units = <<120-90=30>>30 units.} \\
    \textcolor{blue}{\textbf{Plan:}} \\
    \textcolor{blue}{To solve the problem follow these steps:} \\
    \textcolor{blue}{1. Determine the total number of units in the building. We can multiply the number of units per floor by the total number of floors.} \\
    \textcolor{blue}{2. Determine the number of occupied units. We can multiply the total number of units by the fraction representing the occupied units.} \\
    \textcolor{blue}{3. Determine the number of unoccupied units. We can subtract the number of occupied units from the total number of units.} \\
    \textcolor{black}{...} 
  \end{minipage}
}
\captionsetup{type=figure}
\caption{The few-shot prompt $C^{''}$ for GSM8K high-quality plan creating. The black color text is the requirement $R$. The teal and blue color text is the one of example pairs, which contains input $\overline{x_1}$ including example question and solution and the output $\overline{y_1}$ including example plan.}
\label{fig:C_p+s}
\end{figure*}

\end{document}